\documentclass[a4paper, 10pt, conference]{ieeeconf}      
\usepackage{amsmath,amssymb,amsfonts}
\usepackage{algorithmic}
\usepackage{graphicx}
\usepackage{textcomp}
\usepackage{xcolor}

\IEEEoverridecommandlockouts                              
\title{\LARGE \bf
ZS-SLR: Zero-Shot Sign Language Recognition from RGB-D Videos}

\author{Razieh Rastgoo$^{1,2}$, Kourosh Kiani$^1$,Sergio Escalera$^3$\\
$^1$Semnan University ~~$^2$Institute for Research in Fundamental Sciences (IPM)~~\\ $^3$Universitat de Barcelona and Computer Vision
Center
\\
}

\begin{document}

\thispagestyle{empty}
\pagestyle{empty}
\pagestyle{plain}
\maketitle

\begin{abstract}
Sign Language Recognition (SLR) is a challenging research area in computer vision. To tackle the annotation bottleneck in SLR, we formulate the problem of Zero-Shot Sign Language Recognition (ZS-SLR) and propose a two-stream model from two input modalities: RGB and Depth videos. To benefit from the vision Transformer capabilities, we use two vision Transformer models, for human detection and visual features representation. We configure a transformer encoder-decoder architecture, as a fast and accurate human detection model, to overcome the challenges of the current human detection models. Considering the human keypoints, the detected human body is segmented into nine parts. A spatio-temporal representation from human body is obtained using a vision Transformer and a LSTM network. A semantic space maps the visual features to the lingual embedding of the class labels via a Bidirectional Encoder Representations from Transformers (BERT) model. We evaluated the proposed model on four datasets, Montalbano II, MSR Daily Activity 3D, CAD-60, and NTU-60, obtaining state-of-the-art results compared to state-of-the-art ZS-SLR models.
\end{abstract}

\section{INTRODUCTION}
Sign Language Recognition (SLR), as a structured form of human gestures, contains the visual motion of different parts of the human body to transfer information between the deaf and hearing community. While many applications benefit from SLR \cite{b2}, this area is a challenging research area in computer vision. One challenge is regarding the variation of a sign meaning due to the small variation in the hand shape, orientation, movement, hand location, body posture, and non-manual features, such as facial expressions. In addition, viewpoint changes and sign language evolution over time are other challenges in this area. Like other research areas, SLR also faces the challenges of data annotation. This challenge is more clear in the recent SLR models, based on deep learning. While deep learning approaches have obtained state-of-the-art performance in most vision tasks \cite{b2,b6,b7,b00,b1,b3,b4,b5,rbm,bb8,color,face,bb9,bb10}, they require a huge number of labelled training samples. To address the annotation bottleneck in SLR, we explore the idea of recognizing unseen signs without the annotated visual samples by using their textual information. In this way, we propose a Zero-Shot Learning (ZSL) model for Sign Language Recognition from two input modalities. ZSL-based models aim to recognize the unseen classes that are not visible during training. While the traditional models use the same classes during the training and test, ZSL models use separate classes in training and test \cite{Bilge}. 

Generally, there are two main methodologies in ZSL-based tasks: embedding-based and generative-based models. Projecting from the visual features into a semantic attribute space and generating the samples of the unseen classes are the main goals of these methodologies. In this paper, we propose an embedding-based model using Transformer \cite{Transformer-Object}, Long Short Term Memory (LSTM) \cite{LSTM}, and Bidirectional Encoder Representations from Transformers (BERT) \cite{BERT} from two input modalities for ZS-SLR.

Although classical SLR models have obtained high performance \cite{b1,b2,b3,b4,b5}, they do not work effectively with unseen classes. To address this challenge, we propose a two-stream Zero-Shot Sign Language Recognition (ZS-SLR) model from two input modalities. Two vision Transformer models are used for the human detection and visual features representation. To the best of our knowledge, this is the first time that the RGB and Depth modalities are used in a two-stream architecture for ZS-SLR. Relying on the multi-modality and accurate features, the proposed model works effectively on the public datasets even in coping with the unseen classes.

Our main contributions are as follows: (i) We formulate the problem of ZS-SLR with no annotated visual examples and propose a two-stream model, including the vision Transformer model, LSTM network, and BERT model, (ii) We configure a transformer encoder-decoder architecture, as a fast and accurate human detection model, to tackle the challenges of the current human detection models, (iii) We propose the first two-stream model for ZS-SLR. Three modalities, RGB, Depth, and text, are used in the proposed model to benefit from the complementary capabilities of these modalities, (iv) A step-by-step analysis of the proposed model is presented on four datasets using two evaluation protocols. Our model outperforms the state-of-the-art models in ZS-SLR, Zero-Shot Gesture Recognition (ZS-GR), and Zero-Shot Action Recognition (ZS-AR).

The remainder of this paper is organized as follows. Section 2 briefly reviews recent works in the ZS-SLR, ZS-GR, and ZS-AR. The proposed model is presented in detail in section 3. Results are discussed in section 4. Section 5 and 6 analyze the results and conclude the work, respectively.

\section{LITERATURE REVIEW}
In this section, we briefly review recent work in ZS-AR, ZS-GR, and ZS-SLR. 

\textbf{ZS-AR:} Gupta et al. proposed a deep learning-based model for ZS-AR from skeleton data. The generative space of the latent representations is learned using a Variational Auto Encoder (VAE). A fusion of the visual features and the embeddings corresponding to the textual descriptions is used in the model. Results on the NTU-60 and NTU-120 datasets show that the model outperforms the state-of-the-art models in ZS-AR with the 4.34\%\ and 3.16\%\ relative improvement, respectively \cite{b6}. Mishra et al. proposed a probabilistic generative model for ZS-AR by using a Gaussian distribution model for a class representation. The new samples from unseen classes are synthesized by sampling from the class distribution. Results on three datasets, UCF101, HMDB-51, and Olympic, show the comparable performance with the state-of-the-art methods in ZS-AR \cite{Mishra}. Wang and Chen proposed a video-based ZS-AR model by employing the textual descriptions and action-related still images for semantic representation of the human actions. A CNN is used for visual features extraction. Results on UCF101 and HMDB-51datasets confirm the effectiveness of the proposed semantic representations \cite{Wu18}. Gupta et al. proposed a generative-based model, entitled SynSE, for ZS-AR. This model learns progressively to refine the generative embedding spaces of the visual and lingual modalities. Results of SynSE on the NTU-60 and NTU-120 skeleton action datasets show a relative accuracy improvement of 0.92 \%\ and 3.73\%\ compared to state-of-the-art models, respectively \cite{sota-ntu-4}. Schonfeld et al. proposed a generative-based model, namely CADA-VAE, where a shared latent space of image features and class embeddings is learned by modality-specific aligned Variational Auto Encoders (VAEs). During the training, the distributions learned from images and from class labels are aligned to construct the multi-modal latent features. Results on several datasets show that the model obtains state-of-the-art performance in ZS-AR \cite{sota-ntu}. Wray et al proposed a model, entitled JPoSE, for ZS-AR by disentangling Parts-Of-Speech (PoS) in the video captions. A separate multi-modal embedding space is considered for each PoS tag. The outputs of multiple PoS embeddings are input to an integrated multi-modal space for action retrieval. Results on the EPIC and MSR-VTT datasets show the effectiveness of the JPoSE model for ZS-AR \cite{sota-ntu-2}. Hahn et al. presented a model, namely Action2Vec, by incorporating linguistic embedding of the class labels with extracted features from the video inputs. A C3D model is combined with a two-layer LSTM network for spatio-temporal features extraction. Results on the UCF101, HMDB-51, and Kinetics datasets show that the Action2Vec model achieves state-of-the-art in ZS-AR with 1.3\%\,, 4.38\%\,, and 7.75\%\ relative margins, respectively \cite{Hahn}. Bishay et al. designed a model, entitled Temporal Attentive Relation Network (TARN), for ZS-AR. The TRAN model contains a C3D network combined with a Bidirectional Gated Recurrent Unit (Bi-GRU). A single vector is obtained from the embedding module to map into the Word2Vec embedding. Results on the UCF101 and HMDB-51 datasets show a competitive performance of the proposed model compared to the state-of-the-art alternatives in ZS-AR \cite{Bishay}. 

\textbf{ZS-GR:} Madapana et al. proposed a deep learning-based framework for ZS-GR. They combined two datasets, CGD 2013 and MSRC-12, to develop a database of gesture attributes, including a range of categories. A deep learning-based model, including a Convolutional Neural Network (CNN) and a Recurrent Neural Network (RNN), is proposed to optimize the semantic and classification losses. Results show the effectiveness of the proposed model, obtaining a recognition accuracy of 38\%\ for ZS-GR \cite{Mada2,Mada1}. Madapana et al. proposed an adaptive learning paradigm to indicate the amount of transfer learning carried out by the algorithm. In addition, this model aims to show how much knowledge is necessary for an unseen  gesture to be recognized. To this end, they used some user semantic descriptors to improve the performance of the ZS-GR model. Results on own data show the effectiveness of the proposed solution for ZS-GR \cite{Mada3}. 

\textbf{ZS-SLR:} Bilge et al. proposed a model for ZS-SLR using the combination of I3D and BERT. The body and hand regions are used to obtain the visual features through 3D-CNNs. The longer temporal relationships are obtained via Bidirectional Long-Short-Term Memory (BLSTM) network. Relying on the textual descriptions, they extended the current ASL dataset, namely ASL-Text, which includes 250 signs and the corresponding sign descriptions. Results on this dataset show that this model can provide a basis for further exploration of the ZSL in SLR \cite{Bilge}. In this way, we propose a ZS-SLR model from two input modalities.

\section{Proposed model}
\begin{figure*}[ht]
\centerline{\includegraphics[width=0.8\linewidth]{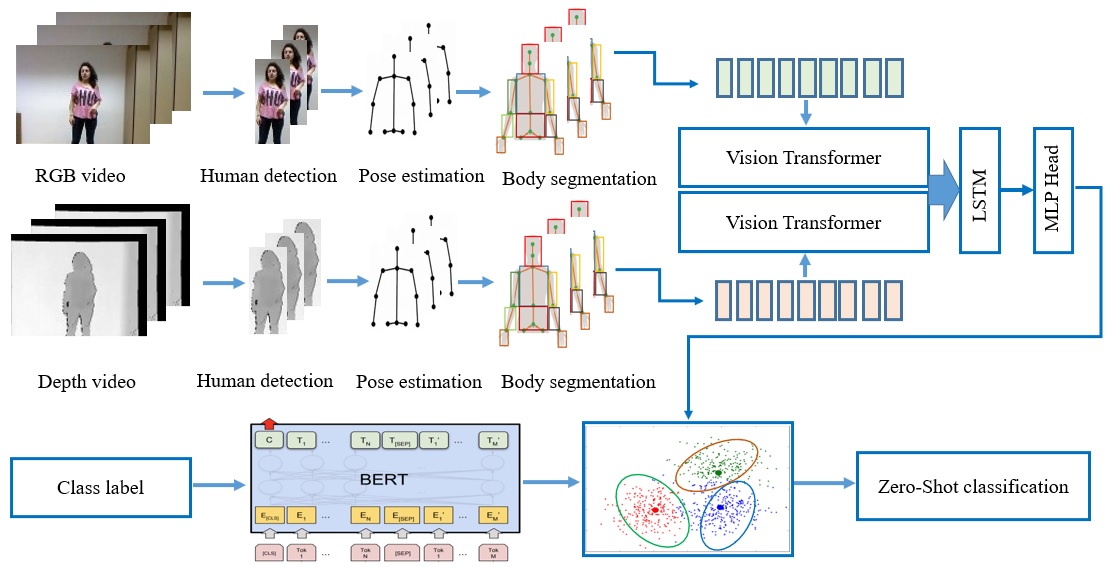}}
\caption{The proposed model including five main steps: Human detection, Pose estimation, Human segmentation, Spatio-temporal visual and lingual features extraction, Zero-Shot classification.}
\label{fig 1}
\end{figure*}

Details of the proposed model are presented in this section. An overview of the model is shown in Fig. \ref{fig 1}.

\subsection{Problem definition}
ZS-SLR uses two sources of information: the visual and lingual domains. While the former domain contains the sign videos, the latter one includes the textual information corresponding to the sign videos. The video signs along with the corresponding class labels are available only for the seen classes. During the test time, the goal is to correctly classify the examples of the unseen classes.

More concretely, let consider the training set $V_{S} = \{(v_{i},c_{i})\}_{i=1}^{N}$, containing N video samples. Each sample includes a video sign ($v_{i}$) and the corresponding class label ($c_{i}$). The model is trained on the training samples. The point is that class labels are not available during the test phase. In this way, the goal is to learn a Zero-Shot classifier that can correctly classify an unseen video sign based on the textual information corresponding to the class label. To this end, given a video sample during the test phase, the model predicts the corresponding semantic embedding using the trained model. The nearest neighbor of the predicted embedding is used for the classification target. Finally, the output of the trained model, F(·), will be as follows:
\begin{equation}
    F(x_{U}^{i}) = \underset{c_{U} \in C_{U}}{\mathrm{argmin} \cos{(f(v_{U}^{i}),BERT(c_{U}^{})})},
\end{equation}
where $\cos$ is the cosine distance and the semantic embedding is computed using BERT \cite{BERT}, $BERT: c_{U}^{} \to R^{1024}$. The function $f$ is a combination of the visual and lingual encoding.

\subsection{Details of the proposed model}
Different blocks of the proposed model are presented in the following.

\subsubsection{Inputs}
Three input modalities, RGB video, Depth video, and text, are used in the model. The visual features are extracted from the RGB and Depth video modalities. The text modality is used for lingual embedding. The visual features and lingual embedding are the input and output of the semantic space, respectively. 

\subsubsection{Human detection}
There are some widely-used and deep learning-based models for object detection, such as Faster-RCNN \cite{F-RCNN}, Single Shot Detector (SSD) \cite{SSD}, and You Only Look Once (YOLO) v3 \cite{YOLO}. However, the detection accuracy of these models is highly related to the hand-based procedures like a Non-Maximum Suppression (NMS) or anchor generation. The explicitly encoding of the prior knowledge about the task is essential in these procedures. To overcome these challenges, we configure a Transformer-based model, namely DEtection TRansformer (DETE) \cite{Transformer-Object}, for human detection. DETE, developed by Facebook AI, contains an end-to-end architecture that employs the Transformer capabilities.

\subsubsection{Human pose estimation}
We use the OpenPose model \cite{OpenPose} for human pose estimation. 13 keypoints are obtained from the human body. We will use these keypoints in the next step.

\subsubsection{Human body segmentation}
\begin{figure}[ht]
\centerline{\includegraphics[height=0.4\linewidth]{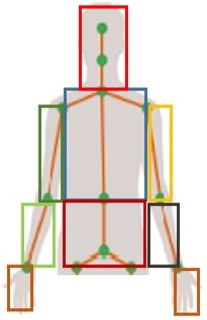}}
\caption{The human body segmentation step in the proposed model. Nine parts are obtained.}
\label{fig 2}
\end{figure}

Considering the skeleton representation of the estimated human keypoints, the human body are segmented into nine sections, as Fig \ref{fig 2} shows. These segments are used to feed to the Transformer model to obtain the visual features representation. 

\subsubsection{Features representation}
We use two feature types in the model: visual and lingual features.

\begin{figure*}[ht]
\centerline{\includegraphics[width=0.8\linewidth]{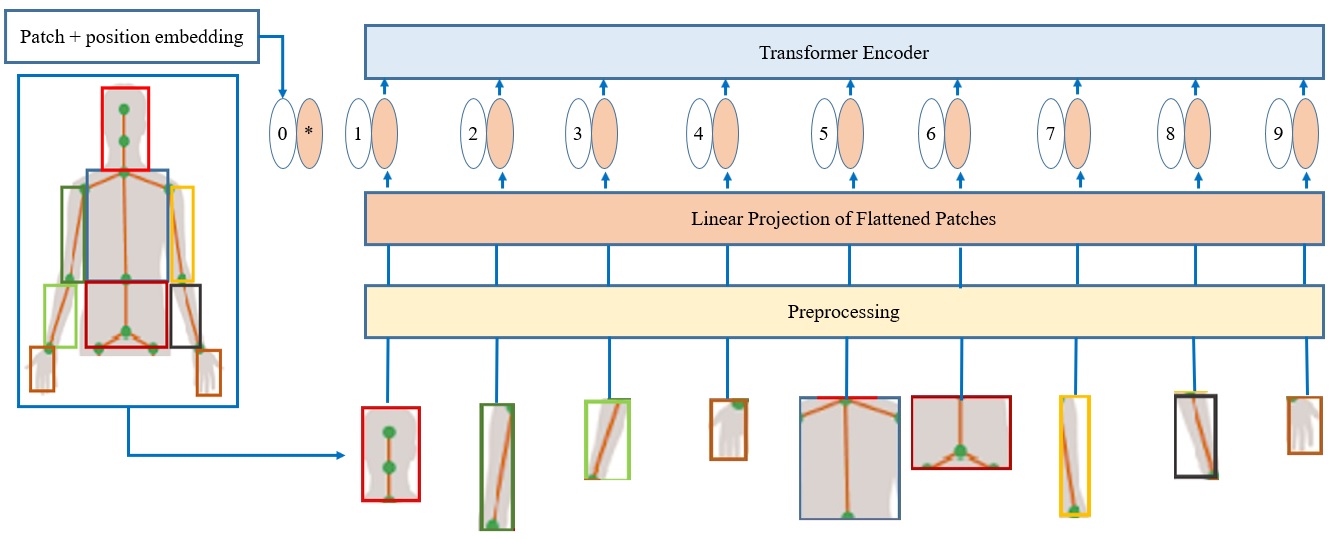}}
\caption{The vision Transformer model used in the proposed model.}
\label{fig 3}
\end{figure*}

\textbf{Visual features:} We use the vision Transformer, developed by Google Brain Team \cite{ViTrans}, to obtain discriminative visual features. To this end, nine body segments are input to the encoder-decoder transformer along with a LSTM network for temporal learning (See Fig. \ref{fig 3}). The human body segments are prepossessed to obtain nine same size segments to input to the vision Transformer model.

\textbf{Lingual features:} We use the sentence BERT model \cite{BERT} to obtain a 1024-dimensional word embedding.

\subsubsection{Features fusion} Two visual features representation obtained from the RGB and Depth modalities are fused to input to the semantic space.

\subsubsection{Semantic space} 
The semantic space aims to map the visual features to the lingual embedding by employing a projection function learned using deep networks. To this end, the similarity degree between the predicted embedding and the unseen class embedding is calculated as follows:
\begin{equation}
    Similarity(c_{i},z_{1})= \cos{(BERT(c_{i}),z_{1})},
\end{equation}
where $c_{1}$ is the class label corresponding to the ith class from the unseen data and $z_{1}$ is the predicted class embedding using the trained projection network. This prediction is compared to all class embeddings obtained from the unseen data to select the closest one.

\subsubsection{Classification} In this step, a Softmax layer is used to recognize the final class label.

\section{Results}
Details of the model implementation and obtained results are presented in the following.

\subsection{Implementation details}
Our evaluation has been done on Intel(R) Xeon(R) CPU E5-2699 (2 processors) with 90GB RAM on Microsoft Windows 10 operating system, Python software, and PyTorch library on 10 NVIDIA Tesla K80 GPUs. Results on four datasets are reported on the unseen test data. The Adam optimization algorithm and a learning rate of 1e-3 are used for 300 epochs. The input resolution is 384X384X3. We select 13 body keypoints to create the body segmentation and feed them to the vision Transformer model for visual feature representation. An LSTM network with 1024 hidden neuron is used for temporal learning. The lingual embedding includes a 1024 embedding vector. A deep model, including two Fully Connected (FC) layers, is used in the semantic space. Two evaluation protocols are used for analyzing the results. These protocols are configured such that we can make a fair comparison with state-of-the-art models.

\textbf{First evaluation protocol:} In this protocol, approximately 90\%\ of the classes is randomly picked up for training and the remaining is reserved for testing. This protocol assigns more samples to the seen data category.

\textbf{Second evaluation protocol:} In this protocol, approximately 70\%\ of the classes is randomly picked up for training and the remaining for testing. This protocol is more challenging compared to the first protocol.

\subsection{Datasets}
Montalbano II \cite{Montal}, MSR Daily Activity 3D \cite{MSR}, CAD-60 \cite{CAD}, and NTU-60 \cite{NTU} datasets are used for model evaluation. Details of these datasets can be found in Table \ref{Table 1}.\\

\begin{table}[h!]
\thispagestyle{empty}
\caption{\label{Table 1} Details of the datasets used for evaluation.}
\begin{center}
{\small
 \noindent\begin{tabular}{p{3.3cm}p{0.7cm}p{2cm}p{1cm}}
 \hline
\textbf{Dataset} & \textbf{Class} & \textbf{Sample Num.} & \textbf{Type} \\
\hline\hline
Montalbano II  & 20  & 12575 & RGB-D\\
MSR Daily Activity 3D  & 16  & 320 & RGB-D\\
CAD-60 & 12 & 60 & RGB-D\\
NTU-60 & 60 & 56'000 & RGB-D\\
\noalign{\smallskip}\hline
\end{tabular}
 }
\end{center}
\end{table}


\subsection{Experimental results}
Results of the proposed model using the two protocols are reported in the following.\\

\subsubsection{Ablation analysis}	
We analyze the effect of different configurations of the proposed model.

\textbf{Different input modalities:} To evaluate the impact of input modalities, we firstly include only one RGB/Depth video modality in the model to benefit from the pixel-level or distance information corresponding to these modalities. After that, we include both modalities in a two-stream architecture to benefit from their complementary capabilities. The deep visual features corresponding to two input modalities are extracted using a compositional model, including two Transformer, LSTM, and BERT model. Results of the proposed model using one/two input modalities are shown in Table \ref{Table 2}. As one can see in this table, the proposed model obtains the highest performance using two input modalities.

\begin{table*}[h!]
\thispagestyle{empty}
\caption{\label{Table 2} Recognition accuracy of the proposed model using different input modalities.}
\begin{center}
{\small
 \noindent\begin{tabular}{p{2.3cm}p{2cm}p{0.8cm}p{2cm}p{0.8cm}p{2cm}p{0.8cm}p{1.5cm}p{0.8cm}}
 \hline
\textbf{Visual Modality} & \multicolumn{2}{c}{\textbf{Montalbano II}} & \multicolumn{2}{c}{\textbf{MSR Daily Activity 3D}} & \multicolumn{2}{c}{\textbf{CAD-60}} & \multicolumn{2}{c}{\textbf{NTU-60}}\\
\hline\hline
 & P1 & P2 & P1 & P2 & P1 & P2 & P1 & P2 \\
\hline\hline
RGB &  50.3 &  43.2 &  49.1 &  41.1 &  49.2 &  43.2 &  59.3 & 54.7 \\
Depth &  40.3 &  36.4 &  42.3 &  36.4 &  38.3 &  40.2 & 46.6 & 46.2 \\
\textbf{RGB and Depth} &  \textbf{62.4} &  \textbf{56.3} &  \textbf{59.2} &  \textbf{52.8} &  \textbf{60.6} &  \textbf{52.5} &  \textbf{79.2} & \textbf{66.4} \\
\noalign{\smallskip}\hline
\end{tabular}
 }
\end{center}
\end{table*}

\textbf{LSTM network:} We evaluate the impact of the number of hidden neurons in the LSTM network in Table \ref{Table 3}. As one can see in this table, the highest performance is obtained using 1024 hidden neurons in the LSTM network.

\begin{table*}[h!]
\thispagestyle{empty}
\caption{\label{Table 3} Recognition accuracy of the proposed model using different hidden layers for LSTM network. In this table, LSTM-N indicates an LSTM network with N hidden neurons.}
\begin{center}
{\small
 \noindent\begin{tabular}{p{2.3cm}p{2cm}p{0.8cm}p{2cm}p{0.8cm}p{2cm}p{0.8cm}p{1.5cm}p{0.8cm}}
 \hline
\textbf{Model} & {\textbf{Montalbano II}} & \multicolumn{2}{c}{\textbf{MSR Daily Activity 3D}} & \multicolumn{2}{c}{\textbf{CAD-60}} & \multicolumn{2}{c}{\textbf{NTU-60}}\\
\hline\hline
& P1 & P2 & P1 & P2 & P1 & P2 & P1 & P2 \\
\hline\hline
LSTM-256 &  58.6 &  52.4 &  53.3 &  50.2 & 56.8 &  49.1 & 66.8 & 65.8 \\
LSTM-512 &  60.2 &  54.2 &  54.1 &  50.9 & 57.4 &  50.5 &  67.2	&  66.0 \\
\textbf{LSTM-1024} & \textbf{62.4} &  \textbf{56.3} &  \textbf{59.2} &  \textbf{52.8} &  \textbf{60.6} &  \textbf{52.5} &  \textbf{79.2} & \textbf{66.4} \\
\noalign{\smallskip}\hline
\end{tabular}
 }
\end{center}
\end{table*}

\textbf{Semantic space:} A semantic space, including two FC layers, is used to project the visual features into the lingual embedding. Analyzing different DNNs showed that a two-layers model had the highest performance (See Table \ref{Table 4}). 

\begin{table*}[h!]
\thispagestyle{empty}
\caption{\label{Table 4} Recognition accuracy of the proposed model using different configurations of the semantic space. In this table, LSTM-N-M indicates an LSTM network with N hidden neurons connected to a DNN  with  M FC layers.}
\begin{center}
{\small
 \noindent\begin{tabular}{p{3cm}p{2cm}p{0.8cm}p{2cm}p{0.8cm}p{2cm}p{0.8cm}p{1.5cm}p{0.8cm}}
 \hline
\textbf{Model} & {\textbf{Montalbano II}} & \multicolumn{2}{c}{\textbf{MSR Daily Activity 3D}} & \multicolumn{2}{c}{\textbf{CAD-60}} & \multicolumn{2}{c}{\textbf{NTU-60}}\\
\hline\hline
& P1 & P2 & P1 & P2 & P1 & P2 & P1 & P2 \\
\hline\hline
LSTM-256-1 &  56.2 &  51.3 &  52.2 &  49.0 &  55.5 & 	48.6 &  66.0 & 64.9 \\
LSTM-256-2 & 58.6 &  52.4 &  53.3 &  50.2 & 56.8 &  49.1 & 66.8 & 65.8\\
Ours (LSTM-512-1) & 59.1 &  53.0 &  53.2 &  49.3 &  56.5 & 	49.6 &  66.4 & 65.1 \\
Ours (LSTM-512-2) &  60.2 &  54.2 &  54.1 &  50.9 & 57.4 &  50.5 &  67.2	&  66.0\\
Ours (LSTM-1024-1) & 61.3 &  55.5 & 58.6 & 51.6 &  59.2 & 51.7 &  69.2 & 65.6 \\
\textbf{Ours (LSTM-1024-2)} & \textbf{62.4} &  \textbf{56.3} &  \textbf{59.2} &  \textbf{52.8} &  \textbf{60.6} &  \textbf{52.5} &  \textbf{79.2} & \textbf{66.4} \\
\noalign{\smallskip}\hline
\end{tabular}
 }
\end{center}
\end{table*}

\textbf{Different human detection models:} Some widely-used object detection models, such as Faster-RCNN, SSD, and YOLO, have been used for human detection. Comparison of the results corresponding to these models with the Transformer model shows that the proposed model achieves a higher accuracy using the Transformer model for human detection (See Table \ref{Table 5}).

\begin{table*}[h!]
\thispagestyle{empty}
\caption{\label{Table 5} Comparison of different human detection models used in the proposed model.}
\begin{center}
{\small
 \noindent\begin{tabular}{p{3cm}p{2cm}p{0.8cm}p{2cm}p{0.8cm}p{2cm}p{0.8cm}p{1.5cm}p{0.8cm}}
 \hline
\textbf{Model} & {\textbf{Montalbano II}} & \multicolumn{2}{c}{\textbf{MSR Daily Activity 3D}} & \multicolumn{2}{c}{\textbf{CAD-60}} & \multicolumn{2}{c}{\textbf{NTU-60}}\\
\hline\hline
& P1 & P2 & P1 & P2 & P1 & P2 & P1 & P2 \\
\hline\hline
Faster-RCNN &  56.1 &  51.6 & 55.2 &  49.8 &  56.6 &  48.4 &  65.6 &  61.4\\
SSD &  58.3 & 53.1 &  58.0 &  51.3 & 58.8 &  50.8 &  67.5 &  63.5 \\
YOLO & 59.6 &  54.2 &  58.4 &  51.9 &  59.4 & 51.2 &  68.1 &  64.2 \\
\textbf{Vision Transformer} & \textbf{62.4} &  \textbf{56.3} &  \textbf{59.2} &  \textbf{52.8} &  \textbf{60.6} &  \textbf{52.5} &  \textbf{79.2} & \textbf{66.4} \\
\noalign{\smallskip}\hline
\end{tabular}
 }
\end{center}
\end{table*}

\subsubsection{Comparison with state-of-the-art models} Using the two evaluation protocols, we compare the results with the state-of-the-art models in ZS-SLR, ZS-GR, and ZS-AR. Results of the proposed model are reported after averaging on ten runs. In each run, we randomly select the training and testing classes. As Table \ref{Table 6} shows, the proposed model outperforms the state-of-the-art models in ZS-SLR, ZS-GR, and ZS-AR.

\begin{table*}[h!]
\thispagestyle{empty}
\caption{\label{Table 6} Comparison with state-of-the-art models}
\begin{center}
{\small
 \noindent\begin{tabular}{p{2.3cm}p{2cm}p{0.8cm}p{2cm}p{0.8cm}p{2cm}p{0.8cm}p{1.5cm}p{0.8cm}}
 \hline
\textbf{Model} & {\textbf{Montalbano II}} & \multicolumn{2}{c}{\textbf{MSR Daily Activity 3D}} & \multicolumn{2}{c}{\textbf{CAD-60}} & \multicolumn{2}{c}{\textbf{NTU-60}}\\
\hline\hline
& P1 & P2 & P1 & P2 & P1 & P2 & P1 & P2 \\
\hline\hline
ZSGL \cite{Mada1} & - & - & 38.1 & - & - & - & - & -\\
ReViSE \cite{sota-ntu-3} & - & - & - & - & - & - & 29.22 & 31.16\\
JPoSE \cite{sota-ntu-2} & - & - & - & - & - & - & 56.49 & 30.75\\
CADA-VAE \cite{sota-ntu} & - & - & - & - & - & - & 65.37 & 35.41\\
GZSL \cite{sota-ntu-4} & - & - & - & - & - & - & 75.81 & -\\
\textbf{Ours} & \textbf{62.4} &  \textbf{56.3} &  \textbf{59.2} &  \textbf{52.8} &  \textbf{60.6} &  \textbf{52.5} &  \textbf{79.2} & \textbf{66.4} \\				
\noalign{\smallskip}\hline
\end{tabular}
 }
\end{center}
\end{table*}

\section{Discussion}
We analyze the proposed model as follows:

\textbf{ZS-SLR:} Recently, deep learning-based models have obtained state-of-the-art performance in SLR \cite{b1,b2,b3,b4,b5,s1,s2}. However, they face the annotation bottleneck and do not work effectively for unseen classes. To tackle the annotation bottleneck, we formulated ZS-SLR with no annotated visual examples and proposed the first two-stream model, including two vision Transformer models, LSTM network, and BERT model. Detailed analysis of the proposed model on four datasets have been performed to provide a basis for further exploration on ZS-SLR.

\textbf{Human detection:} we analyzed some of the widely-used models, such as Faster-RCNN, SSD, and YOLO, for object detection. The performance of these models is highly dependent on the hand-based components, such as a NMS method or anchor production, that require the explicitly encoding the prior knowledge about the task. To overcome these challenges, we configured the DEtection TRansformer (DETE) model for human detection and obtained a higher performance compared to other human detection models.

\textbf{Two-stream modality:} Two visual modalities, RGB and Depth videos, were included in the model to benefit from the complementary capabilities of two modalities. Harnessing from these modalities, the proposed model outperformed the state-of-the-art models in ZS-SLR, ZS-GR, and ZS-AR.

\begin{figure}[htp]
\centerline{\includegraphics[width=0.65\linewidth,height=\textheight,keepaspectratio]{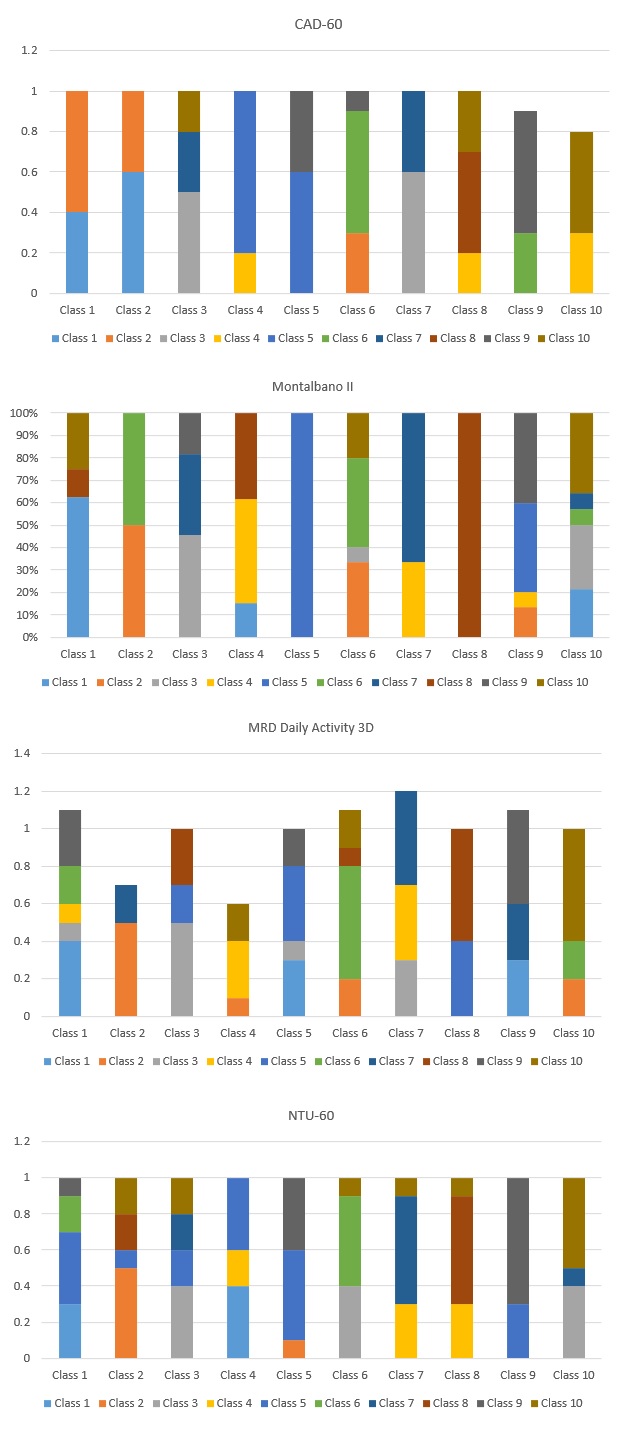}}
\caption{Confusion bars of the proposed model on the four datasets used for evaluation.}
\label{fig 5}
\end{figure}

\begin{figure}[htp]
\centerline{\includegraphics[width=0.8\linewidth,height=\textheight,keepaspectratio]{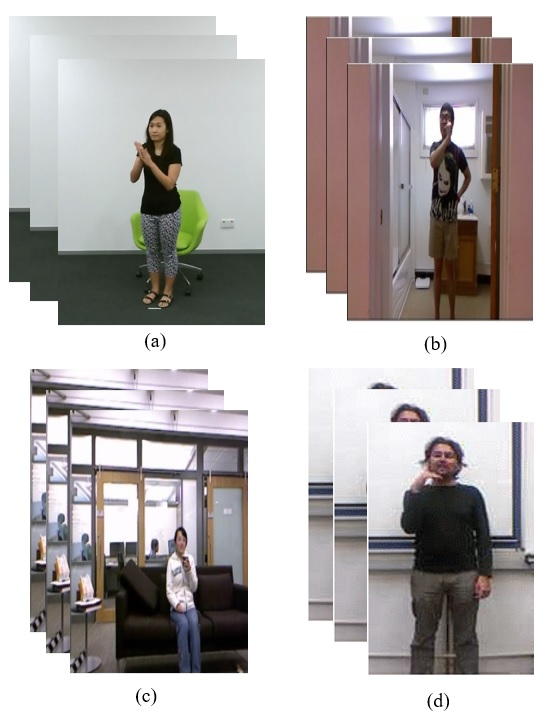}}
\caption{Samples of false recognition of the proposed model on the XX datasets used for evaluation: (a) NTU-60: Predicted: "shake head", Ground truth: "rub two hands together", (b) CAD-60: Predicted: "Rising mouth", Ground truth: "Brushing teeth", (c) MSR Daily Activity 3D: Predicted: "Eat", Ground truth: "Drink", (d) Montalbano II: Predicted: "OK", Ground truth: "Nonme me friega niente".}
\label{fig 6}
\end{figure}

\textbf{Performance:} We performed a step-by-step analysis of the proposed model on four datasets and showed that our model outperformed the state-of-the-art models in ZS-SLR, ZS-GR, and ZS-AR. The false recognition rate was reduced relying on the multi-modality and accurate features obtained from deep blocks. The multi-modality makes the model more robust and effective because the model is not biased to a pre-defined modality. Furthermore, using the vision Transformer capabilities, the proposed model processes the human body segmentation in parallel. While the proposed model outperformed state-of-the-art models in ZS-SLR with high margin, there is much room to improve the recognition accuracy. Detailed analysis on false recognition samples showed that the proposed model obtained a higher performance on the Montalbano II dataset with the recognition accuracy higher than 0.5 in each action. The confusion bars and some samples of false and true recognition can be found in Fig. \ref{fig 5} and Fig. \ref{fig 6}. As these figures show, some signs/actions, such as "shake head", "rub two hands together", "Rising mouth", "Brushing teeth", "Eat", "Drink", "Diving Signals", and "Surgeon Signals", are challenging and difficult to discriminate from the other signs/actions. This comes from high similarities between these signs/actions. As a result, increasing the samples of these actions can decrease the miss-classification error and assist the model to learn complex discriminative patterns. 

\begin{table*}[h]
\thispagestyle{empty}
\caption{\label{Table 7} Class labels per each dataset used for evaluation in confusion matrix.}
\begin{center}
{\small
 \noindent\begin{tabular}{p{1.5cm}p{5cm}p{3cm}p{2cm}p{2cm}}
 \hline
\textbf{Class num.} & \multicolumn{4}{c}{\textbf{Class label}}\\
\hline\hline
& \textbf{Montalbano II}  & \textbf{MSR Daily 3D} & \textbf{CAD-60}  & \textbf{NTU-60}\\
\hline\hline
Class 1	& Vateene & rinsing mouth & drink & brushing teeth\\
Class 2	& Perfetto & brushing teeth & eat & brushing hair\\
Class 3	& E un furbo & wearing contact lense & read book & clapping\\
Class 4	& Nonme me friega niente & call cellphone & reading\\
Class 5	& OK & drinking water & write on a paper & writing\\
Class 6	& Cosa ti farei & opening pill container & use laptop & cross hands in front (say stop)\\
Class 7	& Non ce ne piu & cooking (chopping) & use vacuum cleaner & rub two hands together\\
Class 8	& Ho fame & cooking (stirring) & cheer up & nod head/bow\\
Class 9	& Buonissimo & talking on couch & sit still & shake head\\
Class 10 & Sono stufo & relaxing on couch & toss paper & wipe face\\
\noalign{\smallskip}\hline
\end{tabular}
 }
\end{center}
\end{table*}

\section{Conclusion and future work}
In this work, we proposed a two-stream deep learning-based model for ZS-SLR from two input modalities: RGB and Depth videos. In this model, two vision Transformer models were configured for human detection and visual feature representation. Considering the human keypoints, the detected human body is segmented into nine parts. A spatio-temporal representation from human body is obtained using a vision Transformer and a LSTM network. Finally, the visual features are mapped into the lingual embedding of the class labels, achieved via the BERT model. Results on four datasets, Montalbano II, MSR Daily Activity 3D, CAD-60, and NTU-60, show performance improvement of the proposed model compared to state-of-the-art models in ZS-SLR, ZS-GR, and ZS-AR.\\
 

\addtolength{\textheight}{-3cm}   




\break

\end{document}